# The Humming Revachol: A Dialogue Network Analysis of Narrative Power in *Disco Elysium*

Xinran Liu[1], Miguel Arana-Catania[1], Jack Orchard[2*]

[1] University of Oxford, Oxford, OX1 3BG, UK
[2] Electronic Enlightenment, Osney One Building, Osney Mead, OX2 0EW, UK
* Corresponding author: jack.orchard@bodleian.ox.ac.uk

**Abstract**

This study applies social network analysis to the narrative RPG Disco Elysium to examine how narrative power is distributed among its characters. A directed, weighted character-mention network of 73 NPCs is constructed from the game's complete dialogue script, with characters identified and annotated across five social attributes. Narrative power is operationalised through four dimensions: attention, authority, discursive initiative, and embeddedness. Non-parametric statistical tests examine differences across gender, age, race, class, and ideology. The findings suggest that narrative power is largely independent of social identity categories and is more closely associated with character narrative functions, while structural embeddedness reflects the socio-organisational contexts in which characters are situated. These findings show that dialogue-network analysis can reveal structural patterns of representation that visibility counts alone would miss.

**Keywords***:* Disco Elysium; Social Network Analysis; Character Networks; Game Narratives; Digital Humanities

## Introduction

When players traverse the streets of Martinaise as Detective Harry Du Bois and his partner Kim Kitsuragi, they may well be amazed by how such a small, meticulously crafted district can be populated by such a diverse array of non-player characters (NPCs). These NPCs span age, gender, class, race, and political ideology in an intertwined web of relationships. By showing how power operates through these attributes, *Disco Elysium* constructs an unusually rich social world, prompting the question of how that world is internally organised.

Developed by ZA/UM and released in 2019, this dialogue-driven role-playing game (RPG) casts the player as a detective investigating a murder in Martinaise, a war-scarred district of the fictional city of Revachol. This place is still haunted by a failed communist revolution and the foreign intervention that crushed it half a century earlier (Kłosiński, 2021).

Writer and narrative designer Justin Keenan notes that the game's characters were given distinct functions, ranging from plot facilitators and ideological proxies to figures who mainly contribute to atmosphere, within a world designed to be "deep, rather than broad" (Graft, 2021). Many characters embody overt stereotypes like appearance or accent that call attention to themselves as representations of privilege, class, capital, race, and nationhood, thereby prompting players to examine these underlying ideologies critically (Kjeldgaard-Christiansen & Hejná, 2023). Such function-driven design suggests an uneven allocation of narrative power. The developers' acknowledged left-leaning intellectual commitments (Kiersey & Vazquez, 2025) further suggest that ideological and class positions may carry unequal discursive weight, an expectation tested in this study.

Narrative power, as used here, refers to a character's structural weight within the network of dialogue through which the game's social world is constituted. This is not their moral or political authority within the story, but their prominence, connectedness, and discursive influence relative to other characters. To address this, the present study adopts a two-stage quantitative framework, which is then read alongside close attention to individual characters in the discussion. The first stage employs social network analysis to convert the game's dialogue references into a quantifiable network, translating the abstract notion of narrative power into node-level metrics. Each character here is a node in a network built from who mentions whom in dialogue. A character's structural position refers to where they sit within that network of mentions: how often they are referenced, by whom, and how densely they are connected to other characters. These metrics describe a character's structural position along four dimensions: Attention, Authority, Discursive Initiative, and Embeddedness. Each captures a different mode of being prominent within the network of dialogue relations. The second stage treats these values as dependent variables and tests whether they differ systematically across NPCs' social attributes, using Kruskal–Wallis tests supplemented by Dunn's post-hoc pairwise comparisons. The network analysis thus provides a structural map of narrative power; the statistical testing asks whether that structure is patterned by social identity, and a subsequent close reading of individual characters situates the resulting patterns within the game's narrative content. Together, they allow the distribution of narrative visibility to be examined at a scale that neither approach could address alone.

Building on these contextual, design-based, and methodological rationales, this article raises three research questions. The central question is, how is narrative power distributed across

NPCs within *Disco Elysium*'s dialogue network? Second, to what extent do NPCs' social attributes correspond to differences in their structural position and narrative influence? Third, what do these structural patterns reveal about the broader relationship between authored game design and the symbolic representation of social identity in dialogue-driven interactive narratives?

## Literature Review

Game studies has long examined how narratives encode social power, with the power asymmetry between player characters and non-player characters (NPCs) serving as a central topic. NPCs are traditionally cast as functional figures serving the player's progression and plot advancement (Egenfeldt-Nielsen et al., 2008; Warpefelt, 2016), positioning them as mere agents of action rather than narrative subjects in their own right (Burgess & Jones, 2020). However, recent studies challenge this assumption by showing that game narratives encode social hierarchies of class, race, and ideology through character representation (Fickle, 2019; Murray, 2018; Shaw, 2014). Although frequently using Crenshaw's (1989) intersectionality framework, existing research largely focuses on gender, race, and LGBTQ+ identities while rarely extending to age or ideology (Shaw et al., 2019; Waddell et al., 2014; Williams et al., 2009). Moreover, existing literature heavily emphasises static diversity metrics over how these representations dynamically function and drive narrative power within the game world.

Within this context, *Disco Elysium* serves as a prime case study for studying how games encode power given its themes and narrative design. Existing scholarship across hermeneutics, sociolinguistics, and audience reception confirms the presence of power stratification and ideological conflict in its narrative system. Kjeldgaard-Christiansen and Hejná (2023) use sociolinguistic analysis to show that character voices and accents systematically map onto class status and ideological stance. For instance, Joyce Messier's upper-class British pronunciation and Cuno's working-class Scouse accent serve as key social markers for players. Markić (2024) examines how narrative resources are distributed among the game's four core ideologies (Communism, Moralism, Ultraliberalism, and Fascism), demonstrating that the game establishes distinct dialogue channels and speakers' weights for different ideologies, thereby mirroring real-world imbalances of ideological power. Other researchers who have analysed *Disco Elysium* also offer a wide range of analytical perspectives, reading the game, for example, through Camus's concept of the absurd and the representation of trauma (Spies, 2021), Gothic fiction (Novitz, 2021), posthuman theory (McKeown, 2021), and Deleuze and Guattari's rhizome (Belmonte, 2021), as well as through the political economy of its production and distribution (Gekker & Joseph, 2021).

With advances in computational humanities, character network analysis has become a key framework to unpack narrative power structures. The approach is grounded theoretically in Alex Woloch's (2003) novel character system. Woloch posits that narrative work is fundamentally the allocation of limited narrative attention across an entire character system, where a character's narrative significance is largely defined by their relative position to others. Translating this concept into quantitative structural analysis, Franco Moretti (2011) systematically introduced social network analysis into fictional studies by modelling character interactions in Shakespearean plays as network graphs. By using centrality metrics to quantify how significant or well-connected each character is in the network, Moretti showed that network visualisation reveals patterns of narrative power distribution invisible through close reading. To bridge the gap between qualitative theory and large-scale data, Elson et al. (2010) developed an automated algorithm for extracting dialogue networks from literary texts, making narrative network analysis computationally scalable. Moving beyond literary texts, Jones et al. (2020) applied this framework to *Star Wars* films using directed dialogue networks, quantifying how characters empower or disempower one another through speech to map out narrative power.

Despite its maturity, character network analysis suffers from a media-specific limitation, remaining focused almost entirely on linear media like novels and films. Due to the interactivity hallmark of video games (Weber et al., 2014), player choices exhibit strong path dependence, where early choices restrict future narrative branches. As a result, individual playthroughs capture only a fraction of the game's global text (Aarseth, 1997). This contingency across playthroughs makes standard network models based on single linear texts difficult to apply directly. Computational game narrative research follows two separate trajectories currently, both of which leave key gaps. The first analyses external, derivative texts such as player reviews within online communities (Gabriel, 2025), which captures player reception but fails to reveal the game's internal textual design. The second path extracts in-game dialogue scripts directly for quantitative modelling, but most existing cases are limited to texts with only a small number of branches or to partial slices of a single episode, such as Métrailler's (2023) exploration of narrative flow graphs built from the first chapter of *Life is Strange* or Schlenk et al.'s (2024) script segmentation analysis applied to the relatively linear *Final Fantasy VI*. The most direct precedent is Akoury et al.'s (2023) extraction of *Disco Elysium: The Final Cut*, covering roughly 1.1 million words as a directed graph whose reachability is conditioned by game-state variables and embedded Lua logic. However, the graph serves primarily as training data for masked-utterance infilling, and its topology is never itself interpreted.

Although *Disco Elysium* features a massive volume of text ideal for network analysis, whether dialogue is triggered depends heavily on skill checks and random dice rolls (Belmonte, 2021). This study argues that despite the randomness of player choices, the complete dialogue tree and embedded logic graph preconfigured in the game's underlying code exist as an objective and complete whole within the game text. This project extracted the game's native script data and constructed a full-scale dialogue network across 86 named characters, setting aside the path-dependent contingency of individual playthroughs in order to reconstruct the narrative power landscape as it is encoded by the game's authors. By shifting the analytical focus from single player pathways to the entire underlying dialogue topology, this study provides a generalisable computational model for decoding narrative power in complex, non-linear video games.

## Methodology

### Data Description and Network Construction

The story of *Disco Elysium* takes place almost entirely within Martinaise, a single dilapidated coastal district of Revachol, over the course of roughly seven in-game days. The player controls Harry Du Bois, a police detective who wakes up with near-total amnesia and is tasked, alongside his partner Kim Kitsuragi, with investigating a murder. The murder investigation functions as a narrative spine, but the bulk of the play consists of open-ended exploration and conversation with the district's residents — dockworkers, shopkeepers, political organisers, children, and other figures whose lives intersect with the case only tangentially.

Among the game's roughly ninety named characters, several recur throughout this article. Kim Kitsuragi is Harry's partner and the investigation's methodical anchor. Evrart Claire and Titus Hardie lead rival factions competing for control of the district's docks, the former through formal union politics and the latter through more openly criminal means. Cuno, a foul-mouthed adolescent at the crime scene, and Cunoesse, a separate character whose dialogue largely echoes his, both become entangled in the investigation's early stages. Joyce Messier is an ultraliberal corporate liaison whose upper-class diction contrasts with the Martinaise working-class residents, while Klaasje Amandou is a hostel guest and disco dancer linked to the murder.

The primary dataset comprises dialogue from *Disco Elysium: The Final Cut*, sourced from a community-maintained repository of the game's script data (Disco Elysium Scribe, n.d.). It contains approximately 70,000 rows of dialogue, with each entry recording the speaker, the dialogue text, and the conversation to which it belongs. Only NPC characters were included, and Harry Du Bois's internal voices were excluded. All processed edge lists, annotated character metadata, and analytical code are available in the project repository[1]. Network analysis and statistical testing used Python with NetworkX[2] (Hagberg et al., 2008), SciPy[3] (Virtanen et al., 2020) and the scikit-posthocs[4] (Terpilowski, 2019) for statistical testing.

The network's scope was defined through a manually compiled character profile. Characters were first ranked by in-degree in a preliminary mention network, a directed graph built from all character references in the corpus before filtering. The list was then cross-referenced with community-maintained character databases to add narratively relevant characters, resulting in a final canonical list of 86 named NPCs. Each character was then annotated with five social attributes (gender, age, race, class, and ideology) based on in-game dialogue, character descriptions, and community encyclopaedias (Disco Elysium Wiki, n.d.-a, n.d.-b). Because these encyclopaedias are fan-maintained rather than authorial, the resulting attribute labels, particularly those concerning ideology, should be interpreted with these source limitations in mind.

Character mentions were extracted through string matching using an alias map of 91 name variants, including nicknames, partial names, and spelling differences. The resulting references were aggregated into a directed weighted network in which edge weight represents mention frequency, both direct address and third-person referencing, counted at the level of individual dialogue lines.

To assess the reliability of this approach, a validation was conducted using spaCy named entity recognition (NER) as an independent baseline. Four spaCy English models[5] (small, medium, large, and transformer) were each run on the same corpus with an EntityRuler injecting the canonical name list. All four models detected only 63 unique characters, compared to 86 via string matching, and manual review of over 500 unmatched PERSON entities confirmed that they were fictional place names, objects, or other non-character referents misclassified as persons.

---

[1] https://github.com/Laurel177/disco-elysium-narrative-network
[2] https://networkx.org
[3] https://scipy.org
[4] https://scikit-posthocs.readthedocs.io
[5] https://spacy.io/models/en

Of the 86 canonical characters, 2 neither mention nor are mentioned by any other character, and a further 10 were connected only to Harry, leaving 73.

**Analytic Framework**

Narrative power is understood here as the degree to which a character shapes and occupies the discursive world of the story, represented by their weight within the network of talk through which Revachol's social world is formed. The network is directed: each mention is recorded as a link running from the character who speaks to the character who is spoken about, and repeated mentions increase the weight of that link. Because this weight is not a single property, the study operationalises it across four complementary dimensions, each measured by a distinct network metric. Attention measures a character's overall narrative visibility, defined by the frequency with which they are mentioned by other characters. Authority evaluates the source of that visibility, assessing whether a character is referenced by central, highly discussed figures or by marginal ones. Discursive Initiative captures a character's active agency in the discourse, distinguishing between those who actively direct talk toward others and those who exist merely as objects of conversation. Embeddedness determines a character's position within the network's overall structural layout, measuring whether they belong to a densely interconnected core or remain loosely attached at its periphery. These dimensions are related but not interchangeable. A character may be widely discussed yet never discuss others or occupy a tight structural core while attracting little attention overall.

Attention is measured by weighted in-degree, that is, the total weight of all links pointing toward a character. It captures the most basic form of narrative power, that is who gets talked about. Because repeated mentions by the same speaker are counted, a high in-degree indicates a character to whom the game's dialogue keeps returning, regardless of who is doing the returning. Raw frequency alone, however, cannot distinguish between a character mentioned repeatedly by minor background figures and one mentioned by the story's most central speakers.

Authority addresses this limitation using the authority score from Kleinberg's (1999) HITS algorithm. Consider Klaasje Amandou and Joyce Messier as examples. Klaasje receives 60 mentions to Joyce's 38, yet Joyce scores higher on authority because the characters who talk about Joyce are themselves more central to the network's conversational traffic. Counting mentions alone would place Klaasje above Joyce; weighting those mentions by the prominence of the speakers reverses the ordering. HITS is well suited to directed dialogue networks because it separates two roles: hubs, which mention many others, and authorities, which are mentioned by those hubs. The two scores are calculated together, with hub scores increasing when a character mentions high-scoring authorities and authority scores increasing when a character is mentioned by high-scoring hubs, until the values stabilise. A character therefore receives a higher authority score when the people talking about them are themselves active participants in the network. In narrative terms, this distinguishes a character embedded in the story's dominant discourse from one who is merely mentioned in passing. The hub score is not analysed as a measure of authority; instead, it is used to build the Discursive Initiative measure introduced below, where it captures a character's tendency to direct attention toward others rather than attract it.

Discursive Initiative captures the balance between receiving discourse and directing it. To operationalise this distinction, the study uses the ratio of authority to hub score, here termed the auth–hub ratio (ratio = Authority / Hub). Authority increases when other characters talk about a character (as target), whereas hub score increases when a character talks about many others (as source). The ratio therefore indicates whether a character is primarily the object of others' attention or instead a participant who redirects attention outward. A high value means that the character is more often spoken about than directs discourse toward others; a low value means the opposite. For example, Evrart, with a ratio well above 1, appears largely as an object of others' discourse, whereas Kim actively mentions many characters but is rarely referenced in return, producing a much lower ratio.

While the preceding dimensions capture power as a function of discourse, embeddedness shifts focus to where a character sits within the structure of the network. It is measured by k-core decomposition (Seidman, 1983). Labatut and Bost (2019) note that k-core analysis has been used to distinguish the densely connected core of a novel's cast from its peripheral figures. A k-core is a subgraph in which every character is connected to at least k others inside that same subgraph. The value of k therefore sets the minimum level of connectedness required to remain inside that particular core. Each character is assigned the highest k-core they still belong to, so higher coreness values indicate deeper integration into the denser, more mutually connected regions of the network. Embeddedness thus captures a form of narrative power that visibility alone cannot. A character in the core is bound into the network of mutual reference through which the story's social world is held together, while a peripheral character may be mentioned yet remain structurally detached from it. K-core decomposition operates on undirected graphs, where a connection between two characters records only that a mention occurred between them, without its direction. The directed mention network was therefore symmetrised for this metric alone, treating any pair linked by a mention in either direction as connected.

To interpret the patterns surfaced by the quantitative analysis, selected dialogue passages involving structurally notable characters are examined through close reading.

**Statistical Methods**

Non-parametric tests were used because every metric is strongly right-skewed, with most characters at low values and a few dominant figures forming a long upper tail, and all metrics departed significantly from normality under the Shapiro-Wilk test (all $p < .05$; Shapiro & Wilk, 1965). Parametric alternatives such as ANOVA, which assume roughly normal distributions with comparable variance across groups, would thus yield unreliable estimates. Additionally, Spearman rank correlation coefficients (Spearman, 1904) were computed to evaluate the baseline relationship between coreness and weighted in-degree prior to group-level inferential testing.

For social attributes with three or more groups (age, race, class, and ideology), Kruskal–Wallis H tests were employed (Kruskal & Wallis, 1952) to assess whether groups occupy systematically distinct rank positions in the network hierarchy. For gender, which comprises only two primary groups, the two-sample Mann–Whitney U test was applied (Mann & Whitney, 1947). Because a statistically significant Kruskal–Wallis test indicates only that a difference exists among groups without identifying specific contrasts, significant results ($p < .05$) on HITS-derived metrics were followed by Dunn's post-hoc pairwise comparisons (Dunn, 1964), with Bonferroni correction applied to the resulting p-values to guard against false positives arising from the multiple pairwise tests involved. Characters categorised as "unclear" were omitted from group comparisons, and a minimum subgroup threshold ($n \geq 3$) was mandated for inclusion in statistical testing. All tests were executed using the SciPy and scikit-posthocs (Terpilowski, 2019) scientific computing libraries.

Given the modest subgroup sample sizes and the exploratory aim of mapping where structural differences arise rather than testing pre-registered hypotheses, omnibus tests across social attributes are reported without family-wise correction. Bonferroni correction was applied only within the post-hoc comparisons. Findings are therefore interpreted as indicative patterns warranting further study rather than confirmatory results.

For the Race attribute, two complementary classification schemes were applied, both drawing on the game's own fictional racial categories. Individual attributes were manually extracted from raw script files, cross-referenced against community encyclopaedias (Disco Elysium Wiki, n.d.-a, n.d.-b), and compiled into the NPC attribute dataset. The broad scheme (Occidental, Mesque, and Semenese) aggregates characters based on shared macro-geographic and cultural realms within the fictional world, merging Revacholian and Oranjese characters into "Occidental" due to shared continental alignment, whereas Mesque and Semenese were maintained as distinct groups given their clear physical and cultural differentiation in the text. The specific scheme (Revacholian, Oranjese, Mesque, Semenese, etc.) disaggregates dominant categories into their primary constituent identities. Under this scheme, subgroups below the minimum size were kept in an aggregate "Other" group for network visualisations only, which did not alter test outcomes.

To further assess the robustness of the two significant Kruskal–Wallis results, given the known sensitivity of network metrics to node variation and sampling effects ( Borgatti et al., 2006; Costenbader & Valente, 2003), a leave-one-out (LOO) analysis was conducted. Each member of the relevant attribute groups was removed one at a time, and the Kruskal–Wallis test was re-run to determine whether significance depended disproportionately on any single case. Network metrics were not recalculated after node removal, so the check evaluates whether group-level patterns are driven by influential cases, not whether the network's topology would change under node deletion.

## Results

### Network Overview

Weighted in-degree across the 73 characters is strongly right-skewed (median 19, IQR 5.5–31.5) in Figure 1. Most characters receive fewer than 25 mentions, only 9 exceed 50, and three (Titus Hardie, Evrart Claire, Cuno) exceed 150. Narrative attention is therefore concentrated among a narrow set of figures rather than evenly distributed.

**Figure 1**

*Distribution of Weighted In-degree across 73 NPCs*

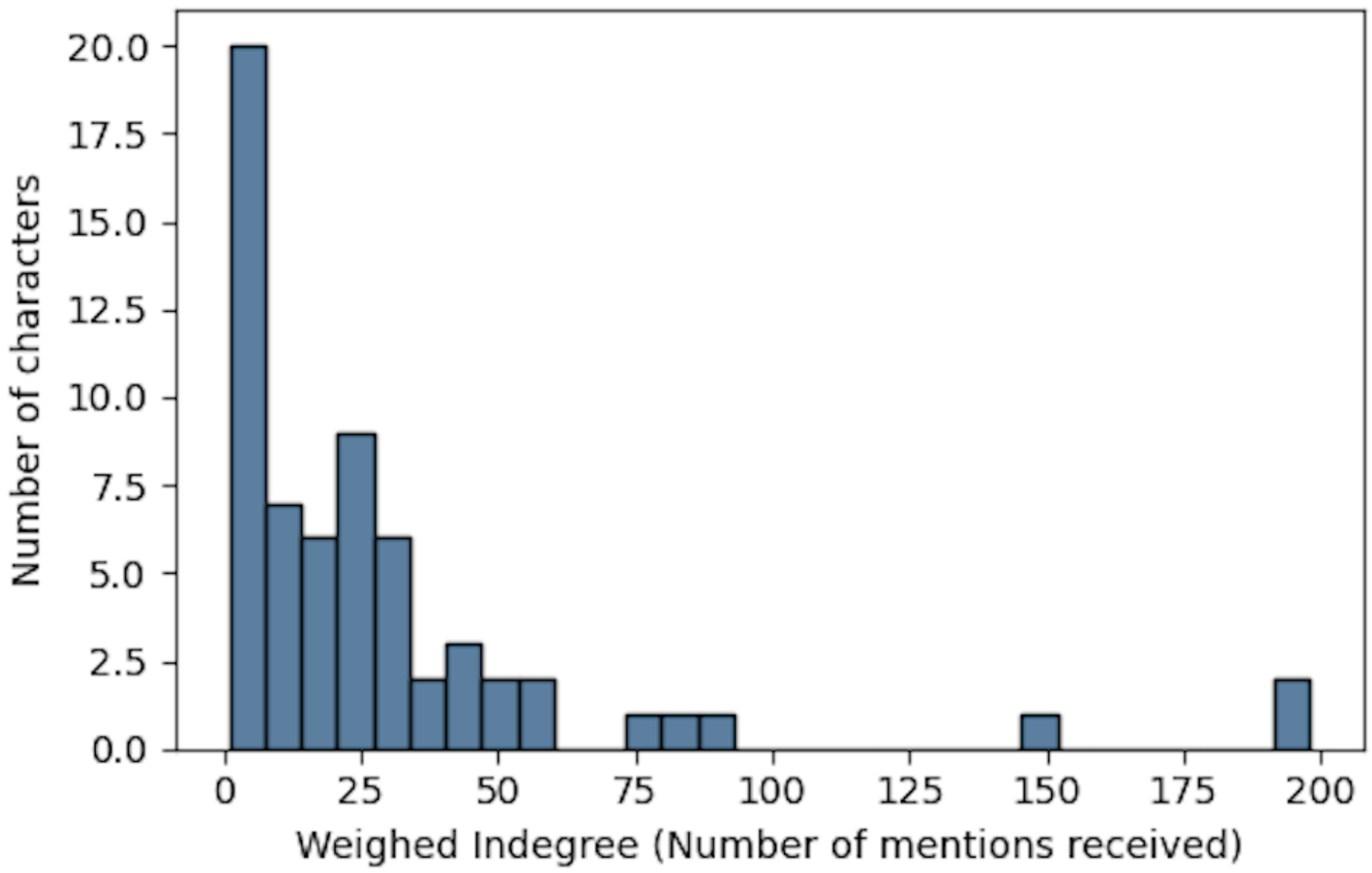

Table 1 presents the full set of group-level comparisons conducted across four narrative power dimensions and five social attributes. Two comparisons reach statistical significance: auth–hub ratio by Race (specific) ($p$ = .041) and coreness by Ideology ($p$ = .042). The remaining sections examine each dimension.

**Table 1**

*Statistical Significance of Group Differences in Narrative Power Metrics Across Social Attributes (Kruskal–Wallis H and Mann–Whitney U tests)*

| Dimension | Metric | Gender | Age | Race (broad) | Race (specific) | Class | Ideology |
|---|---|---|---|---|---|---|---|
| Attention | In-degree | .233 | .051 | .270 | .400 | .627 | .190 |
| Authority | Authority | .742 | .776 | .154 | .148 | .204 | .311 |
| Discursive Initiative | auth–hub Ratio | .993 | .147 | .070 | **.041*** | .818 | .670 |
| Embeddedness | Coreness | .828 | .075 | .059 | .054 | .166 | **.042*** |

*Note.* *$p$ < .05.

## Attention and Authority: Descriptive Patterns

Figure 2 shows the joint distribution of authority and hub scores across all 73 characters, grouped by five social attributes. Most characters cluster near the origin, with a few occupying extreme positions. Overall distribution patterns are consistent across attributes, with no distinct social clustering. Key individual characters anchor the extremes: Cuno has the highest authority score (authority ≈ 0.27, hub ≈ 0.01), being widely discussed but rarely mentioning others. Conversely, Cunoesse holds the highest hub score (authority ≈ 0.00, hub ≈ 0.32), actively referencing others with minimal reciprocal mention. Evrart Claire and Titus Hardie show moderate scores on both axes, while Kim Kitsuragi leans toward a higher hub score, acting primarily as a conversational driver. Statistical tests in Table 1 confirm these observations. Mann–Whitney U tests (Gender) and Kruskal–Wallis H tests (Age, Race, Class, Ideology) showed no significant differences for hub (all $p > .27$) or authority scores (all $p > .14$). Weighted in-degree similarly yielded no significant differences (all $p > .19$), except for Age, which approached significance ($p = .051$).

**Figure 2**

*Authority and Hub Score Distribution*

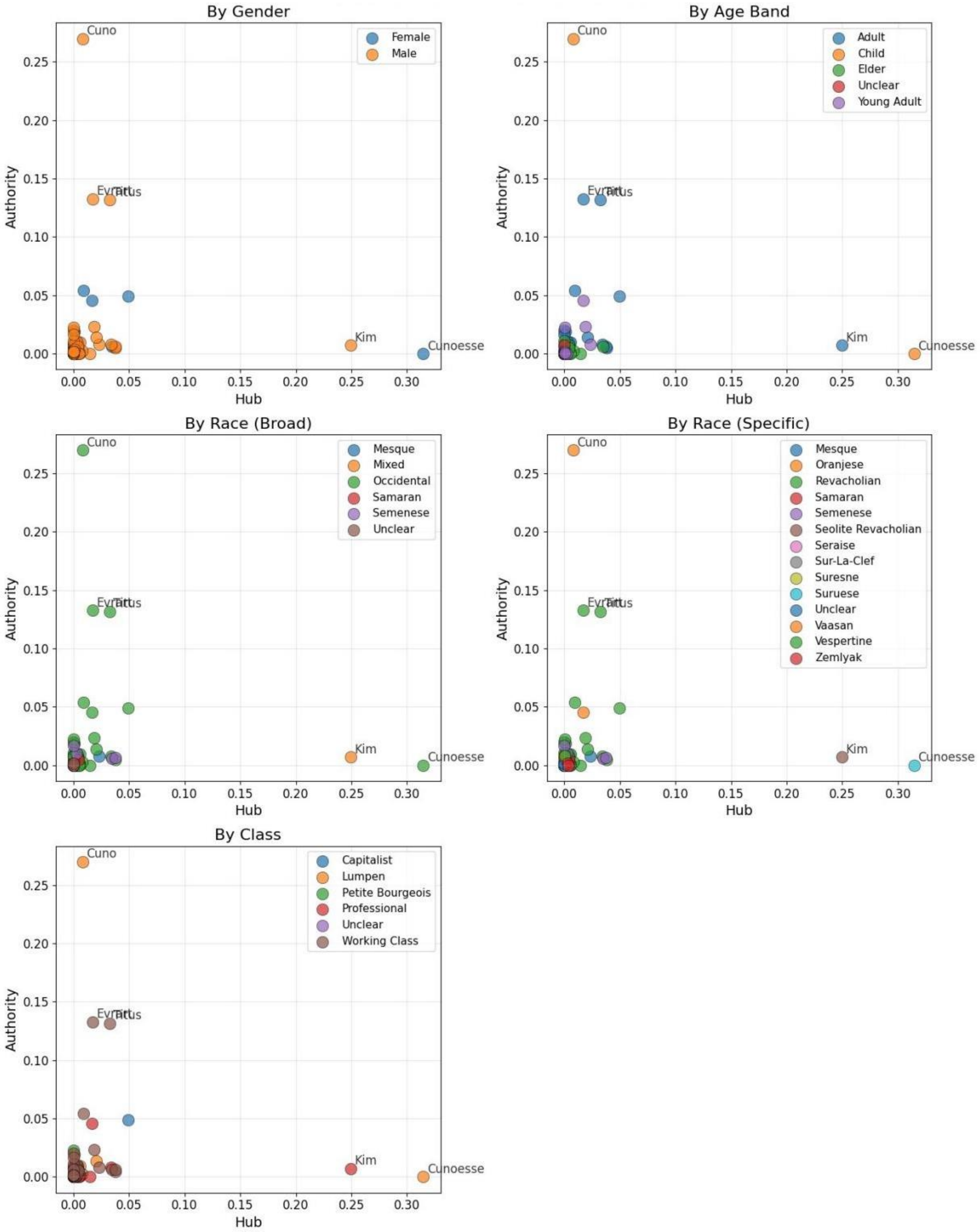

## Discursive Initiative: Race and the Auth–hub Ratio

Among the five social attributes tested, Race (specific) produces the only significant result for auth–hub ratio (Kruskal–Wallis H, $p = .041$), indicating that a character's tendency to be talked about relative to their tendency to talk about others varies across racial groups. Table 2 presents descriptive statistics for each group. By construction, values above 1 indicate that a character is mentioned more often than they mention others. The Oranjese group shows the strongest tendency toward being talked about, the Mesque group the weakest.

**Table 2**

*Descriptive Statistics for Auth–Hub Ratio by Race (Specific) Group*

| Race | n | median | mean |
|---|---|---|---|
| Oranjese | 5 | 2.71 | 14.16 |
| Revacholian | 31 | 1.72 | 7.58 |
| Semenese | 3 | 0.18 | 1.55 |
| Mesque | 5 | 0.00 | 0.55 |

Figure 3 displays the distribution of the auth–hub ratio across the four groups included in statistical testing. Oranjese characters have the highest median ratio (2.71), followed by Revacholian (1.72), Semenese (0.18), and Mesque (0.00). As defined in the operationalisation of the auth–hub ratio, values above 1 indicate that a character is mentioned more than they mention others. The Oranjese group shows the strongest tendency toward being talked about rather than talking, while the Mesque group shows the weakest, with most characters actively mentioning others but receiving little attention in return. Semenese and Revacholian characters fall between these extremes, with Revacholian characters constituting the largest group (n = 31), clustering around a ratio of 1.72. The Race (broad) classification does not reach statistical significance ($p = .070$), suggesting that the observed differentiation is specific to the finer-grained ethnic subgroups rather than broader racial categories.

**Figure 3**

*Auth–hub Ratio Distribution by Race (Specific) Group*

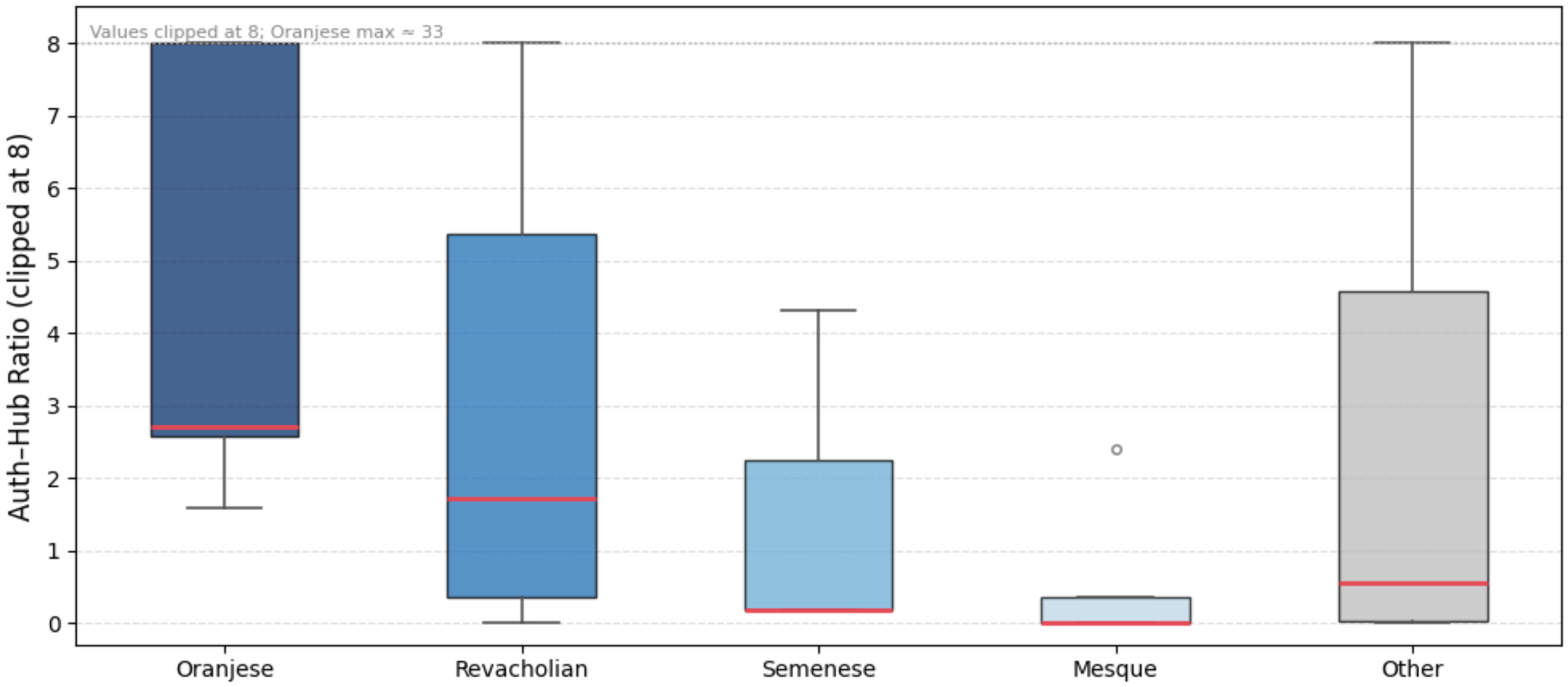

*Note.* Values clipped at 8 for readability. Oranjese maximum = 33.38. Kruskal–Wallis test conducted on Oranjese, Revacholian, Semenese, and Mesque only ($n \geq 3$). Other categories (Suruese, Seolite Revacholian, Vaasan, Vespertine, Zemlyak, Samaran, and Seraise) are shown for visual reference only and excluded from statistical testing.

## Embeddedness: Ideology and Network Position

Figure 4 visualises the network with nodes coloured by coreness level, ranging from purple (coreness 1) at the periphery to red (coreness 6) at the core. The network shows a clear core-periphery structure, with densely interconnected high-coreness characters such as Klaasje Amandou and Kim Kitsuragi at the centre and progressively more peripheral layers beyond. Coreness correlates with weighted in-degree (Spearman $\rho = .737$, $p < .001$) but is not reducible to it: Kim Kitsuragi reaches coreness 6 despite an in-degree of only 35, well below Evrart Claire (192) and Titus Hardie (198), who also reach coreness 6.

**Figure 4**

*K-core network by Coreness Level*

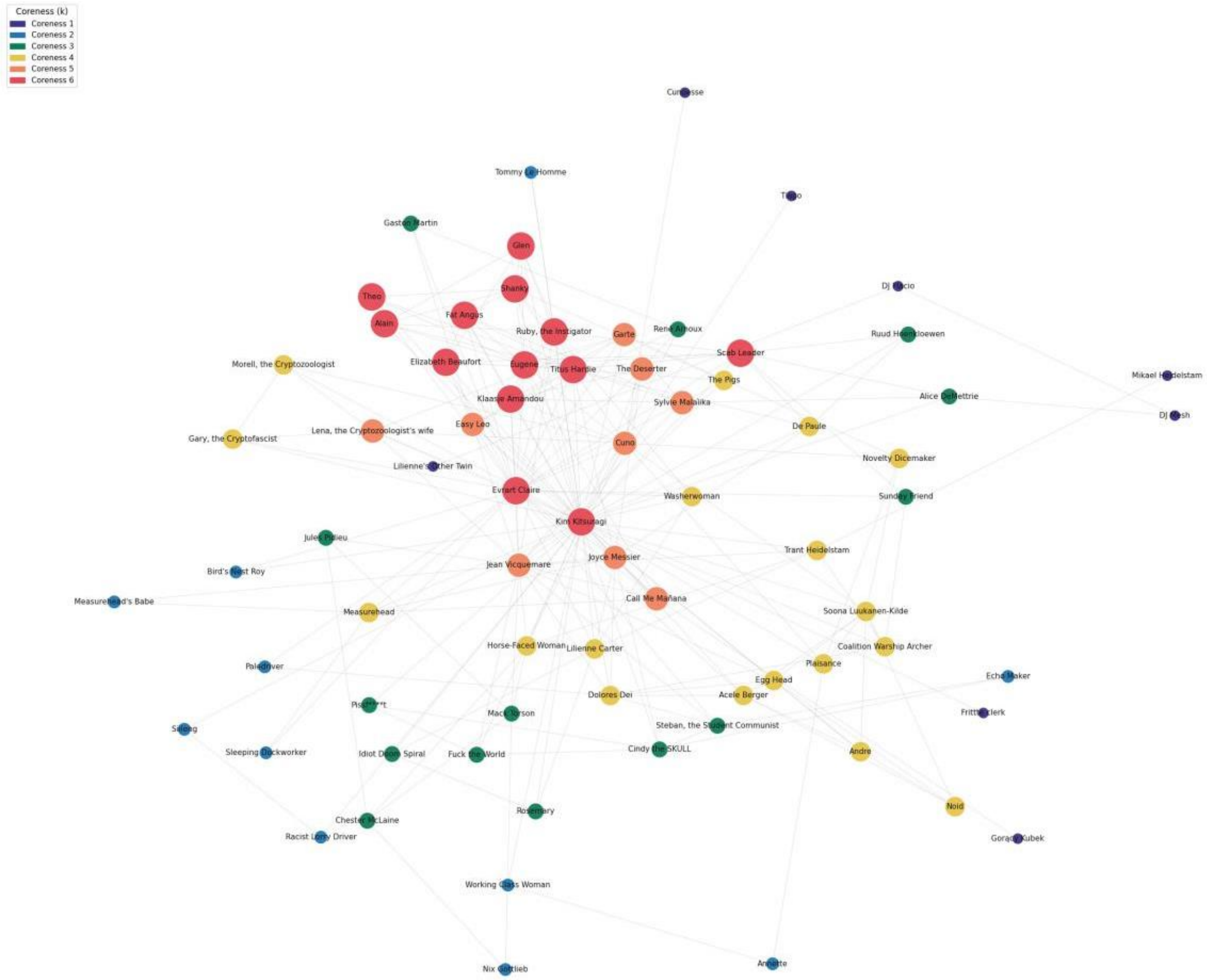

Figure 5 overlays ideological affiliation onto the same network structure, with node size scaled by coreness. Communist characters (blue) are more concentrated in and around the high-coreness core, whereas Ultraliberal characters (yellow) are more peripheral and generally smaller, indicating lower embeddedness. Fascist and Moralist characters are more dispersed across intermediate and peripheral positions, without forming an equally strong single ideological cluster.

**Figure 5**

*K-core Network by Ideological Affiliation*

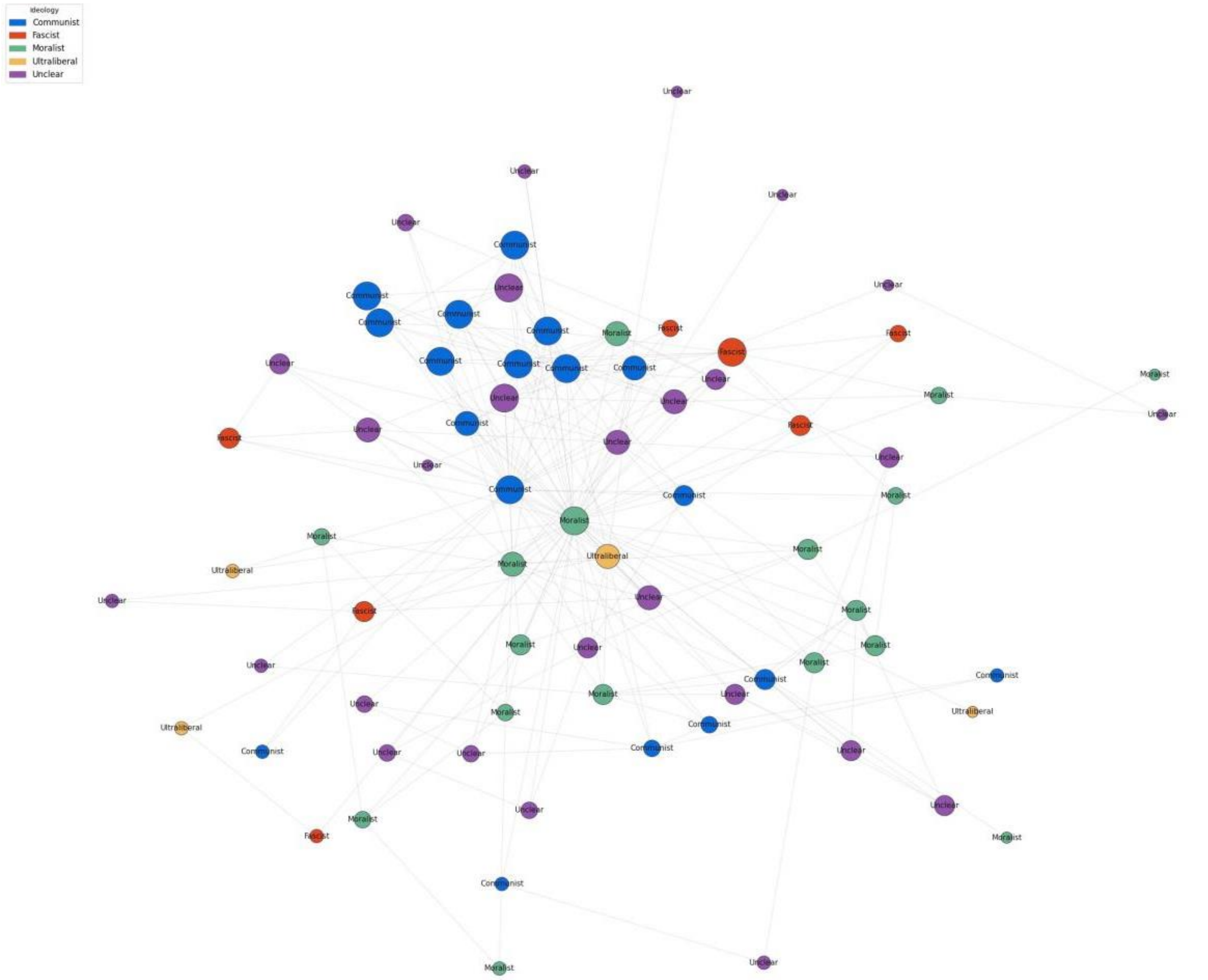

*Note.* Characters coded as "Unclear" are included for visual reference only and excluded from statistical testing.

This visual impression is confirmed by formal testing. A Kruskal–Wallis H test yields a significant difference in coreness across ideology groups ($p$ = .042). Figure 6 displays the distributional details across all four ideology groups. Communist characters have the highest median coreness (5.5), indicating that characters aligned with this ideology tend to occupy the network's denser regions. Ultraliberal characters occupy the opposite end of the distribution, with a median of 2.0. Fascist and Moralist characters cluster at an intermediate level, both with medians of 4.0.

**Figure 6**

*Coreness Distribution by Ideological Group*

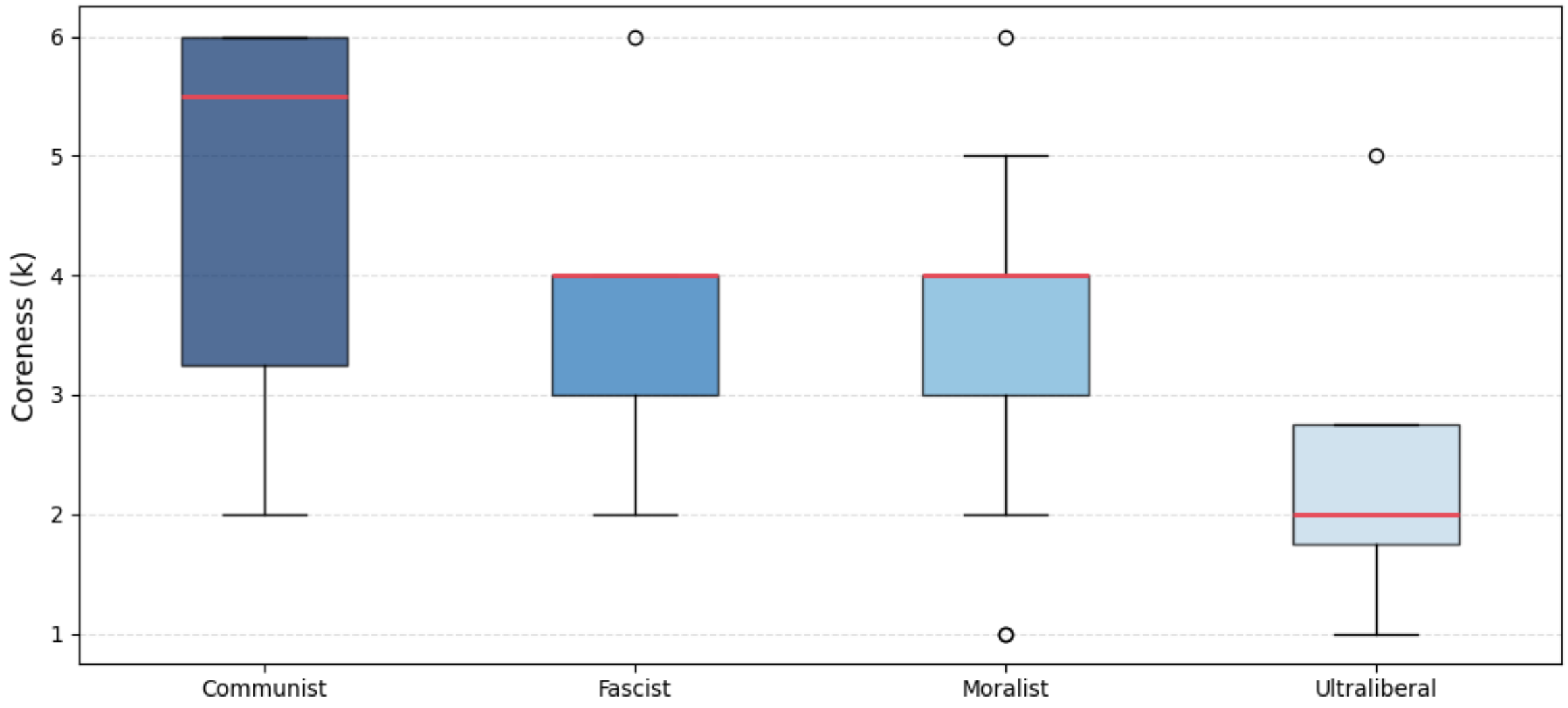

## Summary

Within this overall balance, two localised patterns of differentiation emerge. The auth–hub ratio varies significantly by Race (specific) ($p$ = .041), indicating that ethnic groups differ in the directional orientation of narrative agency regarding whether a character is talked about more than they talk. Coreness varies significantly by Ideology ($p$ = .042), suggesting that ideological affiliation is associated with structural embeddedness within the network core. As both results in Table 1 fall close to $\alpha = 0.05$, LOO sensitivity checks were run (Table 3). The race finding loses significance upon removing either the Oranjese outlier (Cuno, Egg Head) or any Mesque-group member. The ideology finding is more stable, remaining significant under two-thirds of the removals. Both patterns should thus be read as exploratory rather than fully robust.

**Table 3**

*Sensitivity of Key Findings to Single-Node Exclusion*

| Test (target metric) | Full sample (baseline) | Exclusion 1 | Exclusion 2 | Exclusion 3 |
|---|---|---|---|---|
| Race x auth–hub Ratio | H = 8.274, $p$ = .041 | H = 7.322, $p$ = .062 | H = 7.322, $p$ = .062 | H = 6.823, $p$ = .078 |
| Ideology x Coreness | H = 8.219, $p$ = .042 | H = 6.549, $p$ = .088 | H = 6.943, $p$ = .074 | H = 6.943, $p$ = .074 |

The predominant finding across these analyses is structural balance. Weighted in-degree, authority score, and hub score—capturing narrative attention, discursive authority, and conversational initiative, respectively—show no significant differentiation across the five social attributes tested (Gender, Age, Race, Class, and Ideology). Consequently, the concentration of narrative power observed in Figure 1 does not systematically align with social categories, as characters from diverse backgrounds are comparably distributed across visibility and influence levels. This suggests that *Disco Elysium*'s dialogue network distributes foundational narrative prominence in a broadly even-handed manner, while exhibiting specific asymmetries explored in the following discussion.

## Discussion

### Who Is Seen? Attention Without Demographic Hierarchy

Network analysis reveals an extremely right-skewed distribution, indicating that a tiny minority of characters dominates narrative attention. Nevertheless, this concentration does not divide along foundational social attributes such as gender, age, or class. No statistically significant differences were found across either the In-degree or Authority dimensions (all $p > .05$), pointing to a state of structural balance that contrasts with the "visibility bias" documented elsewhere in game studies (Fron et al., 2007; Murray, 2018; Williams et al., 2009), where identity hierarchies from the real world and the games industry are broadly replicated in virtual worlds (Murray, 2018; Shaw, 2014).

In *Disco Elysium*, attention instead tracks a character's function within the case and its locale rather than demographic labels. Before excluding Harry, the highest edge weights belong to his bidirectional exchange with Kim (Kim → Harry 669, Harry → Kim 606), and Kim's next heaviest mentions go to Titus and Evrart (71 and 69), the dockworker and union leaders who serve as key suspects and information sources. After excluding Harry, the single heaviest edge is Cunoesse → Cuno (128). Cunoesse holds the network's highest hub score with an authority score near zero, while Cuno occupies the top authority position, an asymmetry driven by her narrative function as his discursive echo chamber rather than by either character's gender or age.

The absence of significant differences in Authority across all five social attributes (all $p > .05$) indicates that no identity group systematically monopolises discursive authority even when accounting for who is doing the mentioning, not just how often. Klaasje Amandou, for instance, receives 60 mentions against Joyce Messier's 38, yet Joyce holds the higher authority score: as a negotiating representative of transnational capital, she is mentioned by characters who are themselves central to the network, illustrating that authority tracks structural position rather than identity labels.

Equality at the level of visibility, however, is not equality of narrative power, as the next section shows. As Shaw et al. (2019) note, simply "adding in" marginalised groups presumes an absence of structural barriers within the industry while ignoring that the quality and context of representation are far more vital than raw numbers.

**Who Speaks, and Who Is Spoken Of? Asymmetries Beneath the Balance**

The auth–hub ratio can be understood as a character's structural lean between two distinct discursive positions: high-hub characters primarily occupy the productive position of framing and initiating narrative discourse (the Animator/Author), whereas high-authority characters function predominantly as the constructed focus (the Figure) embedded within the talk of others (Goffman, 1981). The race-specific classification scheme exerts a statistically significant effect on the auth–hub ratio ($p = .041$), an effect that disappears under the broad classification scheme ($p = .070$); however, this significance is empirically fragile (full sensitivity results in Table 3).

At one extreme, the Mesque group (median = 0.00) is embedded in daily life as low-wage labourers or marginalised residents (e.g., Paledriver, Novelty Dicemaker, Sleeping Dockworker), with a zero in-degree exemplifying narrative marginality: being "voiced without resonance." At the other extreme, the elevated Oranjese ratio (median = 2.71) is driven by two characters: dropping either Cuno (33.38) or Egg Head (30.57) renders the effect non-significant ($p = .062$). A closer reading of the dialogue shows that the former's mentions are heavily concentrated in a single dialogue partner (Cunoesse), while the latter is strictly localised to the church side quest. In the lore of Martinaise, Oranjese historically signifies foreign capital and occupation order as part of the Coalition forces that suppressed the Revachol Commune fifty years prior (ZA/UM, 2021). However, both Cuno (a local street child) and Egg Head (a dockland moralist) diverge sharply from this colonial authority. There is a clear mismatch between the colonial historical weight carried by the racial tag and the actual character positions driving the statistical pattern. Thus, this race-specific effect should be interpreted as a preliminary observation contingent on localised scene design rather than a robust demographic rule. This mismatch also points to a broader caveat: the racial and ideological classifications used in this study rely on fan-maintained wikis rather than authorial documentation.

In contrast, the alignment between political ideology and topological embeddedness (Ideology x Embeddedness) is comparatively more stable, though it too should be read as exploratory rather than confirmatory (Table 3). The unresolved history of the Revachol Commune's bloody suppression by foreign capital is not mere background lore but acts as a structural force shaping contemporary social relations (Kłosiński, 2021). Under the measure of coreness, Communist characters exhibit deep structural embeddedness. Among the 13 characters occupying the highest coreness tier (coreness = 6), 9 are self-identified Communists, which is a direct outcome of Martinaise's narrative geography as a working-class enclave and its post-colonial history. Rooted in a leftist political ethos of collectivism, solidarity, and organised action, the union network naturally knits Communist characters into a dense, highly interconnected subgraph, embedding their agency within enduring social ties (Granovetter, 1985). Ultraliberal characters present a sharp contrast. Despite their high narrative importance, none occupy the network core (median = 2.00). For instance, Joyce Messier, who commands a high authority score, sits at a structural coreness of only 2. It precisely mirrors her status as an outsider representing the Wild Pines transnational corporation in Martinaise, a position physically epitomised by her Cor-de-Leite '19 luxury sloop docked by the pier. Barred from the harbour and kept apart from the Union's dense ties, she spends her days sipping tea in exasperation over three weeks of enforced inactivity. Other Ultraliberal figures (such as Bird's Nest Roy and the Mega Rich Light-Bending Guy) are similarly dispersed across the network periphery. This structural pattern aligns directly with the capitalist assumption of the "atomised market actor", wherein representatives of capital are extracted from localised social ties and scattered individually across the market (Granovetter, 1985). This structural asymmetry underscores the internalised design of the game's political expression. Ideology and coreness are not fully independent here, since ideology coding partly reflects organisational affiliation (e.g., union, church); the pattern likely reflects their overlap rather than ideology alone.

As lead writer and designer Robert Kurvitz noted, the design team sought to make political expression self-contained within the game's world, refraining from letting "authorial stance" disrupt the text's internal equilibrium (Talbot, 2023). As critic Kunzelman (2019) further observes, characters in Disco Elysium do not appear as abstract individuals who are subsequently tagged with ideology; rather, an NPC's employer, class position, political belief, and living environment are co-constitutive. Ideology possesses a material reality in the game world, and characters live directly within the concrete structural consequences of these institutions.

**Designing Equality on the Surface: Disco Elysium's Narrative Logic**

Synthesising the findings on narrative attention and discursive initiative, the game exhibits balance at the level of "who is seen" (Attention and Authority) such that no identity group systematically monopolises surface-level visibility, a clear distinction emerges on the levels of "who drives the discourse" (Discourse Initiative) and "who occupies the structural core" (Embeddedness). This section argues that the coexistence of surface-level balance and deep-level divergence is the product of a recognisable narrative design logic: narrative power is uncoupled from identity labels and allocated primarily by a character's narrative function, whereas structural position is dictated by the socio-organisational formats in which characters are embedded.

The spatial design of the Martinaise district places the Union at the core of the narrative geography. In the network topology, the Union manifests as a densely interconnected community, whereas foreign capital appears as isolated "islands". This structural pattern is, on its own, consistent with a deliberate representational choice (Shaw, 2014), that the game could easily have placed representatives of capital at the dense core while marginalising the workers, a design choice more typical of mainstream commercial games (Williams et al., 2009), but instead inverted it. The design team's public account of their aims provides further support for interpreting this inversion as an intentional representational choice rather than merely an artefact of the harbour-strike setting.

This framework might seem at odds with the fact that the most central node in the entire network is a white male detective, Harry Du Bois. The network data alone cannot settle whether this amounts to a further instance of centring a white male protagonist or a deliberate subversion of that convention; read alongside the fiction, however, Harry is written as an amnesiac whose personal life has collapsed, with his embarrassment and failure positioned as central to the player's experience (Vella & Cielecka, 2021), which is consistent with the latter.

**Limitations**

There are several limitations of this research. First, the analytical scope was strictly bounded to human NPCs. In *Disco Elysium*, a substantial proportion of dialogue occurs internally between the protagonist, Harry Du Bois, and his 24 psycho-somatic skills, objects, or internal thoughts. By deliberately isolating the external, character-to-character social network to maintain structural consistency, this study does not model Harry's intra-psychic network. Consequently, characters whose narrative presence primarily operates through mental prompts or indirect psychological reflections may be underrepresented.

Second, text mining relies on string matching supplemented by Named Entity Recognition (NER). Literary and ludic scripts frequently employ pronouns, contextual epithets, and indirect references like informal nicknames or fluid titles. Although manual rule-building and wiki cross-validation mitigated false negatives, automated entity extraction inevitably suffers from recall constraints regarding implicit character mentions.

Third, with a localised network size (N = 73), statistical power is necessarily limited, particularly for comparisons involving sparsely populated subgroups (e.g., the in-world national groups Semenese, n = 3, and Oranjese, n = 5). The rarity of these groups follows from the world-building of Revachol, but it also constrains the extent to which the findings can be generalised beyond the present corpus.

Lastly, all attribute categories were annotated by a single coder. Subjective bias was mitigated by adhering to explicit in-game lore, documenting the decisions made for ambiguous edge cases, and cross-referencing community wikis; nevertheless, the absence of a formal inter-coder reliability test means these attribute-level classifications should be interpreted as exploratory rather than definitive.

## Conclusion

This article examined how narrative power is distributed across NPCs in Disco Elysium, whether that distribution varies by social attribute, and what these structural patterns reveal about authored game design and social representation. Using a directed weighted character-mention network of 73 NPCs, measured across four dimensions (Attention, Authority, Discursive Initiative, and Embeddedness), the study finds that narrative power is highly concentrated among individual characters but largely unstratified by social identity. Two exceptions stand out: coreness varies by ideology, and the auth–hub ratio varies by race (specific), with leave-one-out testing showing the latter rests on a small number of cases.

These findings point to a clear logic governing the game's dialogue network. Narrative attention tracks a character's function within the plot rather than their demographic labels; structural embeddedness, by contrast, tracks the forms of social organisation characters are written into, most visibly through union affiliation. Representation in game narrative is therefore not a single property. "Who appears" and "who occupies the structural core" are distinct dimensions that a content analysis of demographic presence alone would not distinguish, and the two can diverge even within a single game. Studies of representation in interactive narrative that rely on visibility counts risk missing exactly the kind of structural asymmetry this analysis identifies.

The method developed here, extracting a game's complete dialogue network directly from its script data rather than from individual playthroughs, makes this kind of structural comparison possible at a scale close reading alone cannot reach, and offers a route other researchers can apply to similarly text-dense, non-linear games.

**Acknowledgements**

We thank the Disco Elysium Scribe project, the disco-reader project and the volunteer editors of the Disco Elysium wikis for providing data.

**Declaration of Conflicting Interests**

The authors declared no potential conflicts of interest with respect to the research, authorship, and/or publication of this article.

**Funding**

The authors received no financial support for the research, authorship, and/or publication of this article.

**Data Availability Statement**

The processed dialogue edge list, annotated character metadata, and analytical code supporting the findings of this study are openly available at https://github.com/Laurel177/disco-elysium-narrative-network